\documentclass{article}

\usepackage{arxiv}

\usepackage[utf8]{inputenc} 
\usepackage[T1]{fontenc}    
\usepackage{hyperref}       
\usepackage{url}            
\usepackage{booktabs}       
\usepackage{amsfonts}       
\usepackage{nicefrac}       
\usepackage{microtype}      
\usepackage{lipsum}		 
\usepackage{graphicx}
\usepackage{natbib}
\usepackage{doi}
\usepackage{array}
\usepackage{ragged2e}
\usepackage{multirow}
\usepackage{longtable}
\usepackage{caption}
\usepackage{float}
\usepackage{graphicx}
\usepackage{subcaption}

\title{What on Earth is AlphaEarth? hierarchical structure and functional interpretability for global land cover
}

\author{ Iván Felipe Benavides-Martínez\thanks{\includegraphics[scale=0.06]{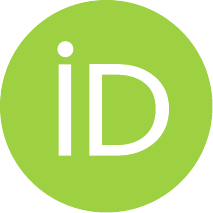}\href{https://orcid.org/0000-0002-1139-3909}{https://orcid.org/0000-0002-1139-3909}} \\
	Artificial Intelligence for Climate and Sustainability\\
	The Institute for Experiential Artificial Intelligence\\
	Northeastern University, Portland, ME, USA \\
	Gulf of Maine Research Institute, Portland, ME, USA \\
	\texttt{i.benavides@northeastern.edu} \\
	\And
	Justin Guthrie\thanks{\includegraphics[scale=0.06]{orcid.pdf}\href{https://orcid.org/0009-0009-9133-6135}{https://orcid.org/0009-0009-9133-6135}} \\
	Sustainability and Data Sciences Laboratory\\
	Northeastern University, Boston, MA, USA \\
	\texttt{j.guthrie@northeastern.edu} \\
	\And
	Jhon Edwin Arias\thanks{\includegraphics[scale=0.06]{orcid.pdf}\href{https://orcid.org/0009-0009-9968-8776}{https://orcid.org/0009-0009-9968-8776}} \\
	School of Engineering and Architecture \\
	Universidad Católica de Manizales\\
	Manizales, Colombia \\
	\texttt{jhon.arias2@ucm.edu.co} \\
	\And
	Yeison Alberto Garcés-Gómez\thanks{\includegraphics[scale=0.06]{orcid.pdf}\href{https://orcid.org/0000-0002-9409-3652}{https://orcid.org/0000-0002-9409-3652}} \\
	School of Engineering and Architecture \\
	Universidad Católica de Manizales\\
	Manizales, Colombia \\
	\texttt{ygarces@ucm.edu.co} \\
	\And
	Angela Ines Guzman-Alvis\thanks{\includegraphics[scale=0.06]{orcid.pdf}\href{https://orcid.org/0000-0002-5185-8950}{https://orcid.org/0000-0002-5185-8950}} \\
	Grupo de Investigación en Recursos Hidrobiológicos \\
	Departamento de Ingeniería \\
	Universidad Nacional de Colombia\\
	Palmira, Colombia \\
	\texttt{aiguzmana@unal.edu.co} \\
	\And
	Cristiam Victoriano Portilla-Cabrera\thanks{\includegraphics[scale=0.06]{orcid.pdf}\href{https://orcid.org/0000-0003-4346-3972}{https://orcid.org/0000-0003-4346-3972}} \\
	Grupo de Investigación en Recursos Hidrobiológicos \\
	Departamento de Ingeniería \\
	Universidad Nacional de Colombia\\
	Palmira, Colombia \\
	\texttt{cvportillac@unal.edu.co} \\
	\And
	Somnath Mondal\thanks{\includegraphics[scale=0.06]{orcid.pdf}\href{https://orcid.org/0000-0001-8217-1361}{https://orcid.org/0000-0001-8217-1361}} \\
	Artificial Intelligence for Climate and Sustainability\\
	The Institute for Experiential Artificial Intelligence\\
	Northeastern University, Portland, ME, USA \\
	\texttt{s.mondal@northeastern.edu} \\
	\And
	Andrew J. Allyn\thanks{\includegraphics[scale=0.06]{orcid.pdf}\href{https://orcid.org/0000-0002-1584-0198}{https://orcid.org/0000-0002-1584-0198}} \\
	Gulf of Maine Research Institute \\
	Portland, ME, USA \\
	\texttt{aallyn@gmri.org} \\
	\And
	Auroop R. Ganguly\thanks{\includegraphics[scale=0.06]{orcid.pdf}\href{https://orcid.org/0000-0002-4292-4856}{https://orcid.org/0000-0002-4292-4856}} \\
	Artificial Intelligence for Climate and Sustainability\\
	The Institute for Experiential Artificial Intelligence\\
	Northeastern University, Portland, ME, USA \\
	\texttt{a.ganguly@northeastern.edu} \\
}

\date{}

\renewcommand{\shorttitle}{\textsc{What on Earth is AlphaEarth?}}

\begin{document}
\maketitle

\begin{abstract}
Geospatial foundation models generate high-dimensional embeddings that achieve strong predictive performance, yet their internal organization remains obscure, limiting their scientific use. Recent interpretability studies relate Google AlphaEarth Foundations (GAEF) embeddings to continuous environmental variables, but it is still unclear whether the embedding space exhibits a functional or hierarchical organization, in which some dimensions act as specialized representations while others encode shared or broader geospatial structure. In this work, we propose a functional interpretability framework that reverse-engineers the role of embedding dimensions by characterizing their contribution to land cover structure from observed classification behavior. The approach combines large-scale experimentation with a structural analysis of embedding–class relationships based on feature importance patterns and progressive ablation. Our results show that embedding dimensions exhibit consistent and non-uniform functional behavior, allowing them to be categorized along a hierarchical functional spectrum: specialist dimensions associated with specific land cover classes, low- and mid-generalist dimensions capturing shared characteristics between classes, and high-generalist dimensions reflecting broader environmental gradients. Critically, we find that accurate land cover classification ($98\%$ of baseline performance) can be achieved using as few as 2 to 12 of the 64 available dimensions, depending on the class. This demonstrates substantial redundancy in the embedding space and offers a pathway toward significant reductions in computational cost. Together, these findings reveal that AlphaEarth embeddings are not only physically informative, but also functionally organized into a hierarchical structure, providing practical guidance for dimension selection in operational classification tasks.
\end{abstract}


\section{Introduction}
\setlength{\parindent}{1em}Geospatial foundation models (GFMs) — part of a broader movement toward comprehensive world models that simulate the dynamics of Earth’s interconnected systems — are transforming Earth observation research in fields such as ecology, infrastructure planning, and climate science \citep{Bodnar2025, Editorial2025}. These models function as "virtual satellites" capable of characterizing the Earth's surface and its dynamics with an unprecedented level of detail \citep{TheAlphaEarthFoundationsteam2025}. GFMs have emerged as a response to two classic bottlenecks in Earth observation: converting large volumes of multi-source data into useful information and the limited generalization of task-specific models when they depend on high-quality labels \citep{Editorial2025}. In this context, Google AlphaEarth Foundations (GAEF) was proposed as a task-agnostic learned featurization approach that integrates multiple data sources, producing eight annual 64-dimensional embeddings at 10 m (2017 to 2024) to support operational mapping and learning with sparse labels \citep{Brown2025, Editorial2025}. 

\begin{figure}
    \centering
    \includegraphics[width=1\linewidth]{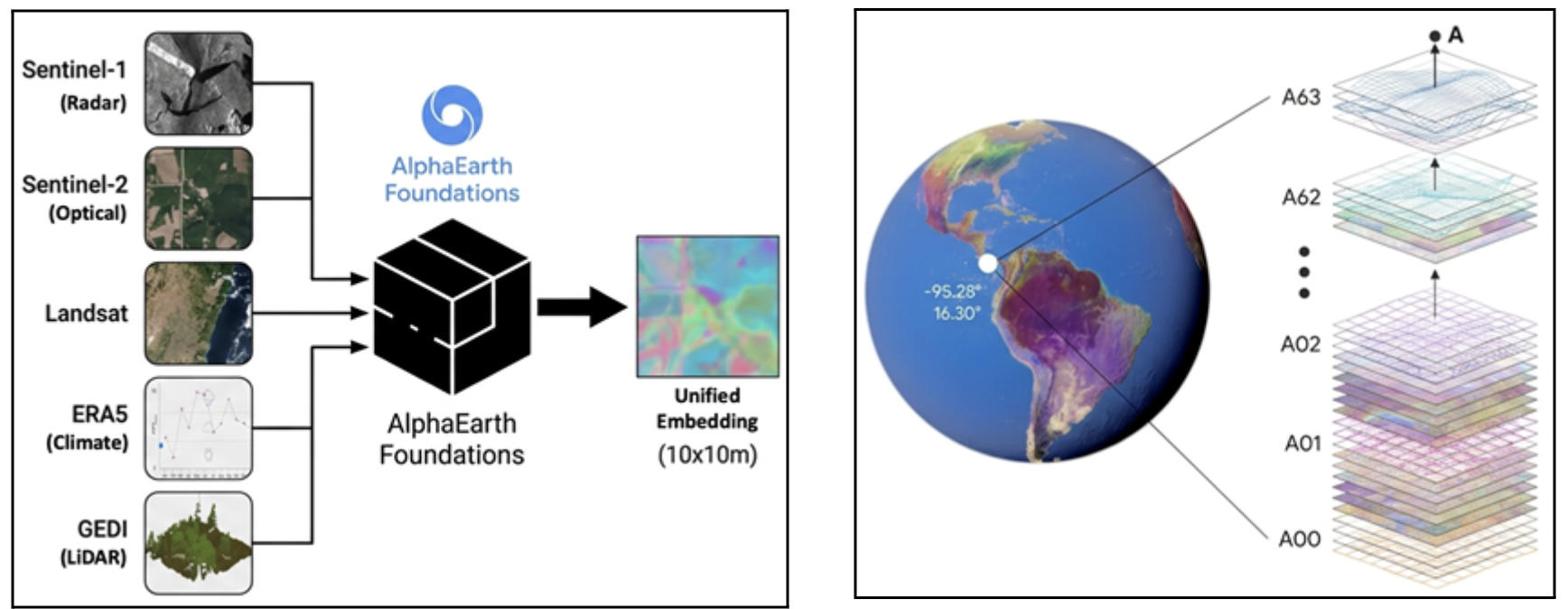}
    \caption{Multi-modal fusion at scale, left) data containing the features, and right) its representation in an embedding.}
    \label{fig:fig1}
\end{figure}

GAEF addresses traditional limitations of remote sensing, such as source heterogeneity, differences in spatial and temporal resolution, and the difficulty of integrating multiple sensors by synthesizing large volumes of data from optical images, radar, LiDAR, climate variables, and other sources (Figure \ref{fig:fig1} left) into a unified and consistent representation of the planet \citep{Houriez2025, TheAlphaEarthFoundationsteam2025}. However, despite demonstrated effectiveness in various mapping and environmental modeling tasks \citep{Brown2025, Ma2025}, these models present a fundamental interpretability challenge: they encode environmental information into high-dimensional embeddings that compress complex geophysical patterns into abstract numerical vectors (Figure \ref{fig:fig1} right) \cite{Brown2025, Houriez2025}. Several criticisms have recently been raised about GAEF, arguing that, despite its strong benchmark performance, it functions as an opaque, annually updated 10‑m ``black‑box'' representation whose mixed‑pixel embeddings, limited physical and human‑centric interpretability, and dependence on biased upstream data constrain its reliability for high‑stakes, fine‑scale, and socio‑ecological decision‑making in real‑world settings \citep{Rahman2026, Liu2025}.

Unlike conventional remote sensing products, where each band is associated with a measurable physical property such as spectral reflectance, temperature, or humidity \citep{TeymoorSeydi2025}, these embeddings represent latent features that capture complex Earth surface patterns without an explicit biogeophysical interpretation \citep{Liu2025}. Their values, usually defined in normalized, high-dimensional spaces, do not offer a direct interpretation of the environmental processes they represent, creating a significant gap between the predictive capacity of the models and their scientific understanding \citep{Rahman2026, Liu2025}. This limitation raises the need to develop methodological frameworks that allow these embeddings to be interpreted in terms of trackable and reproducible physical variables and environmental processes, facilitating their integration into scientific decision-making. Without such frameworks, the utility of embeddings is limited to predictive performance, with little capacity to explain what environmental information is encoded or how they relate to the biogeophysical characteristics of the Earth’s surface. Bridging this gap is particularly relevant for applied domains such as climate adaptation, ecological monitoring, and infrastructure planning, where understanding the structure of model outputs can inform more targeted and efficient use of these representations.

 Furthermore, applied studies explicitly report limitations of GAEF in contexts where interpretability and time sensitivity matter (e.g., agriculture), pointing to low interpretability and other practical constraints such as spatial transferability and limited time sensitivity \citep{Ma2025}. Evidence suggests that GAEF content is rich but poorly characterized: its geometric properties (e.g., normalized embeddings on the unit hypersphere) and its use as a consistent global basis are known, but there is no operational consensus on which dimensions correspond to which variables/processes, nor on how to robustly validate these associations \citep{Liu2025, Rahman2026}.

 Recent studies have begun to show that GAEF embeddings contain physically meaningful information that extends well beyond their original mapping objectives. \citet{Rahman2026} 
 demonstrated that individual embedding dimensions can be associated with continuous environmental variables — such as temperature, vegetation indices, hydrology, and terrain — 
 and that these associations enable large language models to generate interpretable descriptions of land surface conditions with high accuracy. \citet{Qu2026} showed that GAEF embeddings provide richer and more transferable representations of basin characteristics than hand-crafted 
 hydrological attributes, improving streamflow prediction in ungauged basins by capturing integrated environmental signals related to topography, vegetation, and soil properties.  \citet{Koch2025} further demonstrated that GAEF embeddings constitute effective covariates for modelling soil organic carbon and water table depth in peatlands, performing comparably to expert-derived remote sensing inputs while offering substantially lower data preparation 
 barriers.

 However, these approaches have focused primarily on associating embeddings with continuous environmental variables, optimizing their predictive utility for specific downstream tasks, or evaluating their performance against expert-derived covariates — without addressing a more 
 fundamental question: whether the GAEF embedding space itself exhibits a hierarchical functional organization, in which individual dimensions play differentiated roles ranging from 
 high specialization in particular surface conditions to the encoding of broader, shared environmental gradients. Identifying this organizational structure is a prerequisite for systematic, task-independent interpretation of what embeddings encode. Once such a functional chassis is characterized, it becomes possible to derive grounded interpretations of embedding dimensions in relation to the discrete categories that structure the Earth's surface. 
 Among these, land cover stands out as the natural entry point: it is the most immediately perceptible expression of the Earth's surface, globally consistent, spatially pervasive, temporally persistent, and foundational to a wide range of Earth system processes — from carbon cycling and biodiversity to hydrology and climate regulation. Consequently, there is currently no methodological framework that allows the GAEF embedding space to be decomposed into distinct functional roles along a spectrum from specialization to generalization, which limits its use in scientific analysis and knowledge-based decision-making.

 To address this gap, we propose a functional interpretability framework that analyzes the role of embedding dimensions in representing land cover organization. Specifically, we investigate whether the embedding space exhibits a structured, functional organization and whether its dimensions can be systematically categorized according to their contribution to class discrimination and spatial interactions. Our central hypothesis is that embedding dimensions encode differentiated functional roles rather than homogeneous information, and that land cover classes are characterized not by individual dimensions in isolation, but by structured interactions among them. To test this, we combine large-scale experimentation with structural analysis to characterize how embedding dimensions jointly relate to land cover classes.

 The contributions of this work are fourfold: (i) a functional interpretability framework for geospatial embedding dimensions targeting global land cover, (ii) empirical evidence of non-uniform functional behavior across embedding dimensions, (iii) a taxonomy of specialist and generalist embedding dimensions along a functional spectrum, and (iv) a conceptual link between latent representations and geographic structures such as core regions and transition zones.\\

\section{Materials and Methods}
\subsection{Overall study design}
This study proposes a functional interpretability framework for GAEF based on a combination of massive experimentation and structural analysis of the embedding space. The approach is based on the use of embeddings generated by GAEF, which are high-dimensional representations that integrate spatial, temporal, and multi-modal information about the Earth's surface into a unified latent space \citep{Brown2025}. 

 The proposed methodology is structured around two complementary components: a massive experimental exploration aimed at empirically characterizing the discriminative behavior of embeddings; and a structural analysis based on feature importance patterns and progressive ablation that allows interpretations of the functional organization of the representation space in terms of land cover types and their interactions. These components are visualized with an interactive \href{https://alpha-earth-viz.vercel.app/}{Dashboard} that enables exploratory analysis of the embedding space structure and its relationship to land cover classes \citep{Guthrie2025}.
 
\begin{figure}
    \centering
    \fbox{\includegraphics[width=1\linewidth]{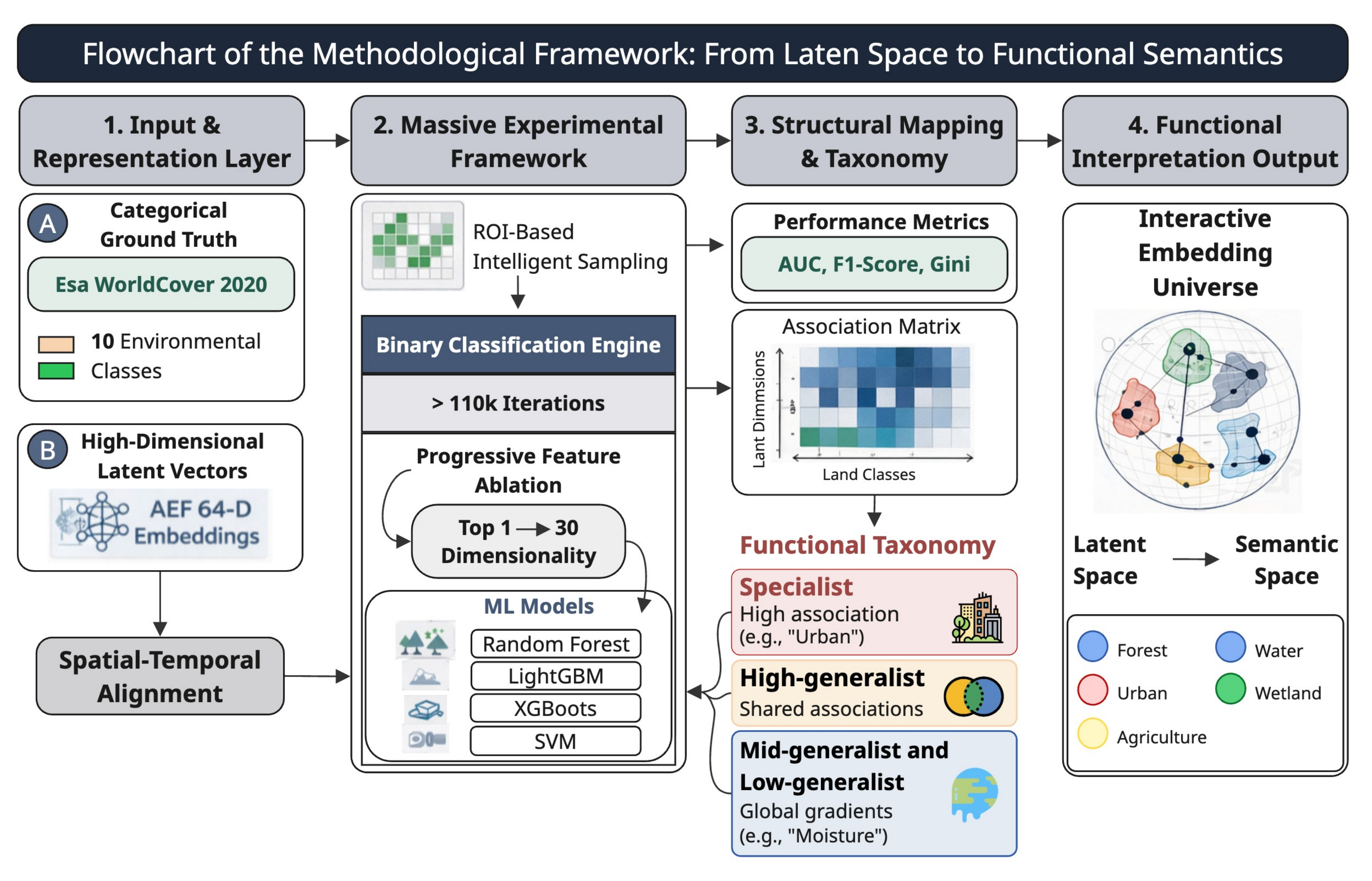}}
    \caption{Flowchart of the methodological framework: from embeddings to functional semantics. The pipeline integrates (1) ESA WorldCover 2020 categorical labels and AEF 64-dimensional embeddings as inputs, (2) a massive experimental framework executing more than 130,000 binary classification experiments using Random Forest, Gradient Boosting, XGBoost, and LightGBM with progressive feature ablation across the top 1 to 30 embedding dimensions, (3) structural mapping via an association matrix and functional taxonomy, and (4) an interactive embedding universe visualization connecting the latent space to semantic space.}
    \label{fig:fig2}
\end{figure}

 This design seeks to transcend evaluation based exclusively on performance metrics, incorporating an interpretive perspective that allows the embedding space to be broken down into differentiated semantic functions, thus addressing the interpretability limitations reported in GAEF applications (Figure \ref{fig:fig2}).

\subsection{Data and representations}

The analysis is based on the integration of two main sources of information: land cover labels and the 64 GAEF embeddings. The labels come from the European Space Agency's (ESA) WorldCover 2020 product, which provides a global classification of the Earth's surface into 11 discrete categories \citep{Zanaga2021}. ESA WorldCover was selected for several reasons: its land cover classes are globally generalizable, providing a consistent categorical framework across diverse geographic contexts; its high spatial resolution (10 m) supports fine-grained discrimination between classes; and its temporal alignment with the period over which GAEF was constructed ensures that the land cover labels correspond to the same conditions encoded in the embeddings. Crucially, ESA WorldCover is independent from the training data used in the construction of GAEF embeddings, meaning that the classification experiments evaluate the embeddings against an external reference rather than recovering labels already encoded during training. These classes represent the categorical structure of the Earth's surface and constitute the target variable in the classification experiments.

The explanatory variables used are the GAEF embeddings, which correspond to the 64-dimensional vectors produced from the integration of multiple observation sources, including optical sensors, radar, climate variables, and other Earth observation products \citep{Brown2025, Houriez2025}.

\subsection{Massive experimental exploration of embeddings}

The first methodological component consists of a large-scale automated experiment designed to evaluate the discriminatory capacity of embeddings in land cover classification tasks. This experiment involves the execution of more than 130,000 independent analyses, each structured as a binary classification in which a target land cover class is evaluated against an aggregation of all remaining 10 classes, thereby isolating the embedding dimensions most informative for distinguishing each individual land cover type. To explore the fundamentals of experimental design and run experiments 
independently, refer to the 
\href{https://mybinder.org/v2/gh/FelipeBenavidesMz/Alpha-Earth-Land-Cover-Classifier/main?labpath=alpha_earth_app.ipynb}{GAEF Land Cover Classification App} 
and to the 
\href{https://github.com/FelipeBenavidesMz/AlphaEarth-Interpretability-Experiments}{Python pipeline} 
developed to run the full set of experiments \citep{Benavides2025, BenavidesMz2025}.

To ensure that each experiment contains sufficient representation of the target class, a class-presence-guided spatial selection procedure is implemented using Google Earth Engine. For each analysis, a continent is randomly selected from a global set and the ESA WorldCover layer is queried to identify a location where the target land cover class is present. Then a rectangular region of interest (ROI) of randomized dimensions ($0.1° - 1.0°$) is constructed around this location. This approach ensures that each ROI contains pixels of the target class while preserving spatial variability across experiments so that the global diversity of land cover conditions is represented. In cases where no target class pixels are found within the selected continent, the system falls back to a random ROI within the continental bounds.

Once the analysis region has been defined, the land cover labels and embedding dimensions corresponding to the selected pixels are extracted simultaneously via stratified sampling, ensuring balanced representation of the target class and all other classes. These data are used to build training and validation sets, in a split of $75/25$, allowing the discriminative ability of each embedding to be evaluated between classes. For each experiment, one of four machine learning algorithms is randomly selected: Random Forest, Scikit-Learn's Gradient Boosted Trees, XGBoost, and LightGBM. All four are suitable for capturing nonlinear relationships and have proven effective in remote sensing classification tasks.

Each experiment follows a training and evaluation scheme in which a model is first trained using all 64 embedding dimensions, and the relative contribution of each dimension to the classification is estimated using the native feature importance scores of the trained model, based on Mean 
Decrease in Impurity (MDI) across decision tree splits.  Based on this ranking, a progressive ablation procedure is implemented in which models are sequentially retrained using subsets of the embedding dimensions, incrementally expanding from the single most important dimension up to the top 30, as ordered by their MDI-based importance scores. This approach allows for the analysis of the redundancy and complementarity of the information contained in the embeddings, as well as the evaluation of how many dimensions are necessary to achieve adequate performance. For each land cover class, the tipping point is defined as the minimum number of dimensions at which mean classification performance across all experiments reaches $98\%$ of the baseline performance obtained with all 64 dimensions (Figure~\ref{fig:fig5}). The dimensions ranked within this tipping point constitute the minimum subset for that class.

The results of each experiment — including performance metrics, variable importance, geographic location, and algorithm used — are systematically recorded in a structured database. This database constitutes the main input for subsequent interpretive analysis, providing large-scale empirical evidence on the behavior of these embeddings.

\subsection{Structural analysis of the embedding space}
The second methodological component focuses on interpreting the embedding space based on the experimental results. While the previous phase allowed us to identify the relevance of each dimension in specific tasks, this phase seeks to characterize the overall organization of the representation space in relation to land cover types.

\begin{figure}
    \centering
    \fbox{\includegraphics[width=1\linewidth]{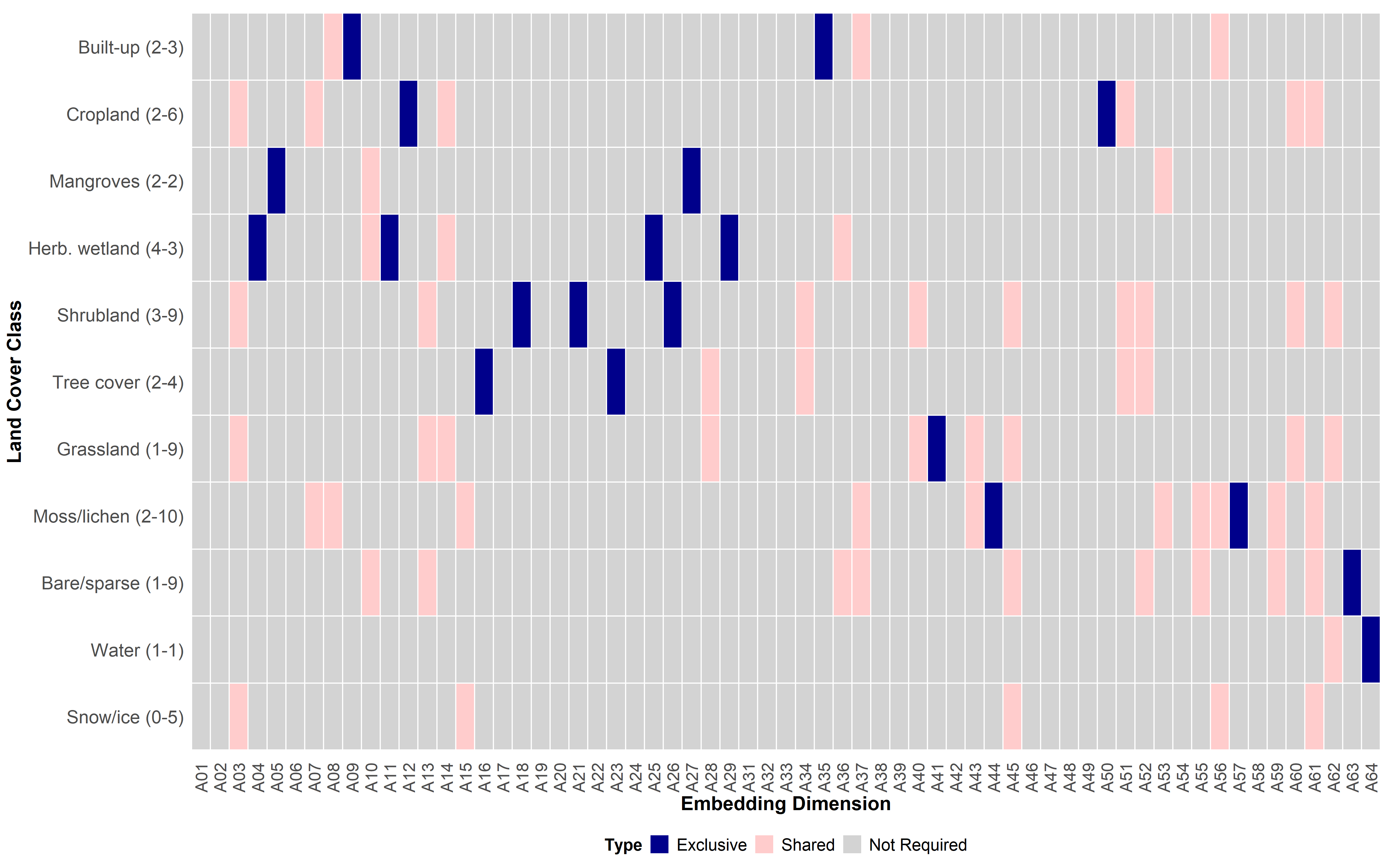}}
    \caption{\textit{Embedding Fingerprint} plot illustrating the functional dimensional signature of each land cover class at the $98\%$ performance recovery threshold. Analogous to a genomic fingerprint, where the presence and absence of bands at specific loci encodes biological identity, each row represents the characteristic pattern of exclusive (blue) and shared (pink) embedding dimensions that together define the discriminative identity of a land cover class within the AEF embedding space. Exclusive dimensions appear in only one class's minimum subset, while shared dimensions --- encompassing low-, mid-, and high-generalists --- contribute to the classification of two or more classes. Gray positions indicate dimensions not required to achieve the $98\%$ threshold for that class. Land cover classes are ordered from top to bottom by increasing spectral complexity, quantified as the mean pixel-wise variance across Sentinel-2 spectral bands computed over 100 randomly selected images per class after standard atmospheric and radiometric corrections. Spectrally homogeneous classes such as Snow/ice and Water --- which exhibit low inter-pixel variability due to their uniform reflectance signatures --- appear at the bottom, while spectrally heterogeneous classes such as Built-up --- whose radiometric response varies substantially across rooftops, roads, and impervious surfaces --- appear at the top, reflecting the increasing number of embedding 
dimensions required to characterize them.}
    \label{fig:fig3}
\end{figure}

To do this, we construct an association matrix that quantifies the relationship between each embedding dimension and the different land cover classes (Table~\ref{tab:land_cover_imp}). For each land cover 
class and each embedding dimension, the association score is computed as the normalized frequency with which that dimension ranked among the two most important features by MDI across all experiments targeting that class. Formally, the association score for a given class--dimension 
pair is defined as the number of experiments in which the dimension appeared among the top two MDI-ranked features, divided by the total number of experiments for that class, yielding a value between 0 and 1. This normalization ensures comparability across classes with different numbers of experiments. The resulting matrix provides a quantitative summary of the discriminative relevance of each embedding dimension for each land cover class, and serves 
as the basis for the fingerprint plot (Figure~\ref{fig:fig3}), the conceptual embedding space representation (Figure~\ref{fig:fig4}), and the embedding universe visualization (Figure~\ref{fig:fig6}) implemented in the interactive Dashboard \citep{Guthrie2025}.

\begin{table}[H]
    \caption{Excerpt of the association matrix quantifying the normalized frequency with which each embedding dimension ranked among the two most important features by MDI across all experiments targeting each land cover class. Values range from 0 to 1, where higher values indicate stronger discriminative relevance of that dimension for the corresponding class}
    \centering
    \renewcommand{\arraystretch}{1.3}
    \setlength{\tabcolsep}{6pt}
    \begin{tabular}{l c c c c c c}
        \toprule
        & \textbf{impA01} & \textbf{impA02} & \textbf{impA03} & \textbf{impA04} & $\cdots$ & \textbf{impA64} \\
        \midrule
        Bare/sparse vegetation  & 0.0267 & 0.0223 & 0.0350 & 0.0167 & $\cdots$ & 0.0409 \\
        Built-up                & 0.0044 & 0.1092 & 0.0478 & 0.0057 & $\cdots$ & 0.0040 \\
        Cropland                & 0.0163 & 0.0266 & 0.0564 & 0.0369 & $\cdots$ & 0.0179 \\
        Grassland               & 0.0211 & 0.0315 & 0.0552 & 0.0270 & $\cdots$ & 0.0161 \\
        Herbaceous wetland      & 0.0177 & 0.0449 & 0.0269 & 0.1090 & $\cdots$ & 0.0199 \\
        Mangroves               & 0.0446 & 0.0103 & 0.0112 & 0.0162 & $\cdots$ & 0.0093 \\
        Moss/lichen             & 0.0093 & 0.0265 & 0.0338 & 0.0136 & $\cdots$ & 0.0285 \\
        Shrubland               & 0.0140 & 0.0278 & 0.0892 & 0.0178 & $\cdots$ & 0.0190 \\
        Snow/ice                & 0.0043 & 0.0041 & 0.1583 & 0.0778 & $\cdots$ & 0.0063 \\
        Tree cover              & 0.0314 & 0.0208 & 0.0454 & 0.0248 & $\cdots$ & 0.0315 \\
        Permanent water bodies ("Water")  & 0.0381 & 0.0983 & 0.0659 & 0.0180 & $\cdots$ & 0.1386 \\
        \bottomrule
    \end{tabular}
    \label{tab:land_cover_imp}
\end{table}

To classify embedding dimensions into functional categories, we leverage the progressive ablation procedure described in Section 2.3. For each land cover class, the progressive ablation procedure identifies the minimum subset of dimensions required to recover $98\%$ of baseline 
classification performance. This threshold was selected because, at this level of performance recovery, the remaining $2\%$ difference falls within the inherent uncertainty of the ESA WorldCover 2020 reference labels themselves, meaning that further gains in classification performance would 
not meaningfully improve agreement with the ground truth beyond what the reference product's own mapping accuracy permits. A dimension is considered ``associated'' with a land cover class if it appears within that class's minimum subset. The number of classes with which an embedding is associated then determines its functional classification: dimensions appearing in only one class's minimum subset are interpreted as \textit{specialists}, while those appearing in the minimum subsets of two, three, or four or more classes are classified as \textit{low-}, \textit{mid-}, and \textit{high-generalists} respectively. For convenience, all non-specialist dimensions are collectively referred to as \textit{shared dimensions}. Dimensions that do not appear in the minimum subset of any land cover class remain uninterpreted under this framework and are discussed in Section 4.6.

This classification approach grounds the functional taxonomy in demonstrated classification performance rather than an arbitrary threshold on importance values, ensuring that each dimension's role is defined by its empirical contribution to accurate land cover discrimination. This approach is supported by evidence that GAEF embeddings capture complex physical and environmental patterns, integrating multiple sources of information into a coherent representation \citep{Brown2025}, and that these representations can reflect relationships with environmental variables at different scales \citep{Rahman2026}.

Additionally, Figure~\ref{fig:fig3} presents what we term the 
``Embedding Fingerprint'' plot --- a visualization whose structure deliberately echoes that of a genomic fingerprint, where the presence and absence of band patterns at specific loci encode the identity of a biological sample. In our case, the on/off pattern of exclusive (blue) and shared (pink) embedding dimensions across positions A01--A64 encodes 
the functional identity of each land cover class, much as specific band combinations in a DNA fingerprint encode the identity of an organism. Just as no two species share an identical banding pattern, each land cover class exhibits a distinctive dimensional signature: specialist dimensions act as unique genetic markers exclusive to a single class, 
while shared dimensions --- encompassing low-, mid-, and high-generalists --- function as conserved sequences that appear across multiple classes, reflecting structural or spectral similarities between them. This visualization provides a compact summary of the dimensional composition 
required to achieve $98\%$ classification accuracy for each land cover type, highlighting differences in the balance between specialist and shared dimensions across classes and offering an intuitive entry point into the functional organization of the GAEF embedding space.

\subsection{Visualization and interpretation of the universe of embeddings}
\begin{figure}
    \centering
    \includegraphics[width=1\linewidth]{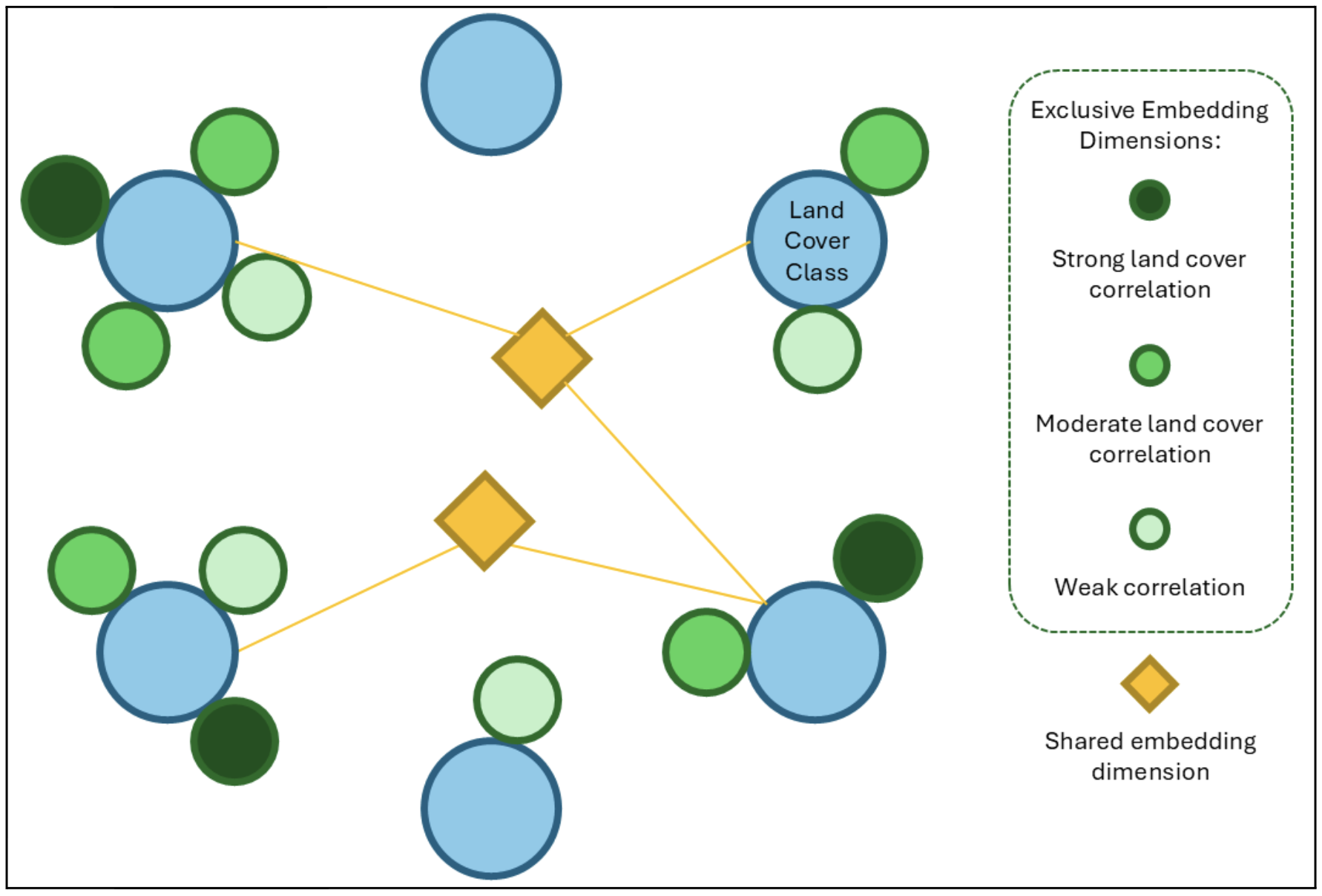}
    \caption{Conceptual structure of the embedding space representation, illustrating the planet-moon organization of the embedding universe. Blue circles represent land cover classes (planets), while green circles represent exclusive embedding dimensions (moons) whose color intensity encodes association strength --- dark green indicating strong discriminative relevance, medium green indicating moderate relevance, and light green indicating weak relevance for that class. Gold diamonds represent shared embedding dimensions that connect multiple land cover classes, positioned at the geometric centroid of their associated classes and reflecting ecotonal or spectrally ambiguous regions where land cover boundaries overlap. This conceptual layout serves as the interpretive basis for the embedding universe visualization implemented in the interactive dashboard \citep{Guthrie2025} and shown in Figure~\ref{fig:fig6}.}
    \label{fig:fig4}
\end{figure}

To facilitate interpretation of the embedding space structure, we developed the \href{https://alpha-earth-viz.vercel.app/}{\textit{``What on Earth is AlphaEarth?''}} interactive dashboard, which visualizes the relationships 
between embedding dimensions and land cover classes in a unified graphical environment \citep{Guthrie2025}. The dashboard integrates classification outputs, embedding importance metrics, and geographic context, enabling 
users to explore the functional organization of the embedding space (Figures~\ref{fig:fig4} and~\ref{fig:fig6}). In its primary view, land cover classes are organized as main nodes arranged in a circular space, while embedding dimensions are represented as entities associated with these classes.

Embedding dimensions classified as specialists are visualized as elements close to a single coverage, reflecting their exclusive association. The darkness of each embedding dimension represents the strength of its association with a particular land cover class, where a darker green indicates a stronger association and a lighter green indicates a weaker one. In contrast, shared embedding dimensions --- those associated with multiple land cover classes --- are located in intermediate positions between their associated coverages, reflecting their role in capturing characteristics common to more than one class. Notably, such shared dimensions may also encode information relevant to ecological transition zones, where land cover types co-occur with one another. This spatial arrangement is calculated using geometric centroids, positioning each embedding dimension according to its relationships with the land cover classes it characterizes.

The Dashboard allows intuitive exploration of the structure of the embedding space, identifying patterns of specialization, clustering, and connectivity between land cover classes. This approach facilitates the interpretation of complex results and provides visual evidence of the functional organization of the embeddings.

The Dashboard comprises four integrated views. The Overview provides a synthesis of the global experiments, including model performance comparisons across the four algorithms employed and a global map of experiment locations. The Class Analysis view houses the ``Embedding Universe'' visualization (Section 3.6), class performance matrices (Section 3.2), embedding importance charts that enable direct comparison of dimensional contributions across classes (Sections 3.2, 3.3). The Geographic view maps individual experiment bounding boxes onto satellite imagery alongside spatially aggregated performance heatmaps, enabling identification of geographic regions where the embedding space performs strongest or weakest (Supplementary Figure~\ref{fig:s2}). Finally, a Chat interface serves as a companion to the experimental notebook (Section 2.3), guiding users through the configuration of new experiments.

\subsection{Methodological integration and hypothesis validation}
The interpretation of each embedding dimension follows a structured inferential process that differs between specialist and shared dimensions. Specialist dimensions are interpreted directly through their exclusive association with a single land cover class: because they contribute 
discriminatively to only one class, their functional meaning is anchored to the distinctive biophysical properties of that class --- such as the permanent water absorption signature of Permanent Water Bodies (embedding A64) or the artificial material reflectance of Built-up areas (embeddings A09 and A35). Shared dimensions, however, require a more deliberate interpretive process. For each shared dimension, we first ask whether a conceptual connection exists between the associated land cover classes that is independent of their geographic distribution --- that is, whether the classes share a common spectral, structural, phenological, or ecological property that could plausibly be captured by a single embedding dimension. If such a conceptual link emerges, it is adopted as the primary interpretation. When no immediate conceptual connection is apparent, we turn to geographic evidence, examining the spatial co-occurrence of the associated classes through maps, satellite imagery, and published literature to determine whether their association reflects a geographically grounded environmental relationship. This two-stage interpretive approach --- prioritizing conceptual parsimony before geographic contingency --- ensures that the functional meanings assigned to shared dimensions are grounded in 
either biophysical logic or empirically observable spatial patterns, rather than in arbitrary statistical co-occurrence.\\

\section{Results}
\subsection{Empirical evidence of the informational efficiency of embeddings}

The results of the large-scale experimental exploration reveal that the discriminatory capacity of GAEF embeddings is not uniformly distributed in the representation space. A small subset of dimensions concentrates information necessary to differentiate land cover types, and models trained with a limited number of prioritized dimensions achieve performance levels comparable to those obtained with the full set of 64 embeddings (Figure~\ref{fig:fig5}a,b).

Across the 11 ESA WorldCover classes, the minimum number of embedding dimensions required to recover 98\% of baseline classification performance varies substantially (Figure~\ref{fig:fig5}c): Water requires only two dimensions, Mangroves four, and Built-up five, while classes with less distinctive spectral or structural signatures --- such as Moss/lichen and Shrubland --- require up to 12 dimensions.

{
\begin{figure}[htbp]
    \centering
    \begin{subfigure}{0.45\textwidth}
        \centering
        \fbox{\includegraphics[width=\linewidth]{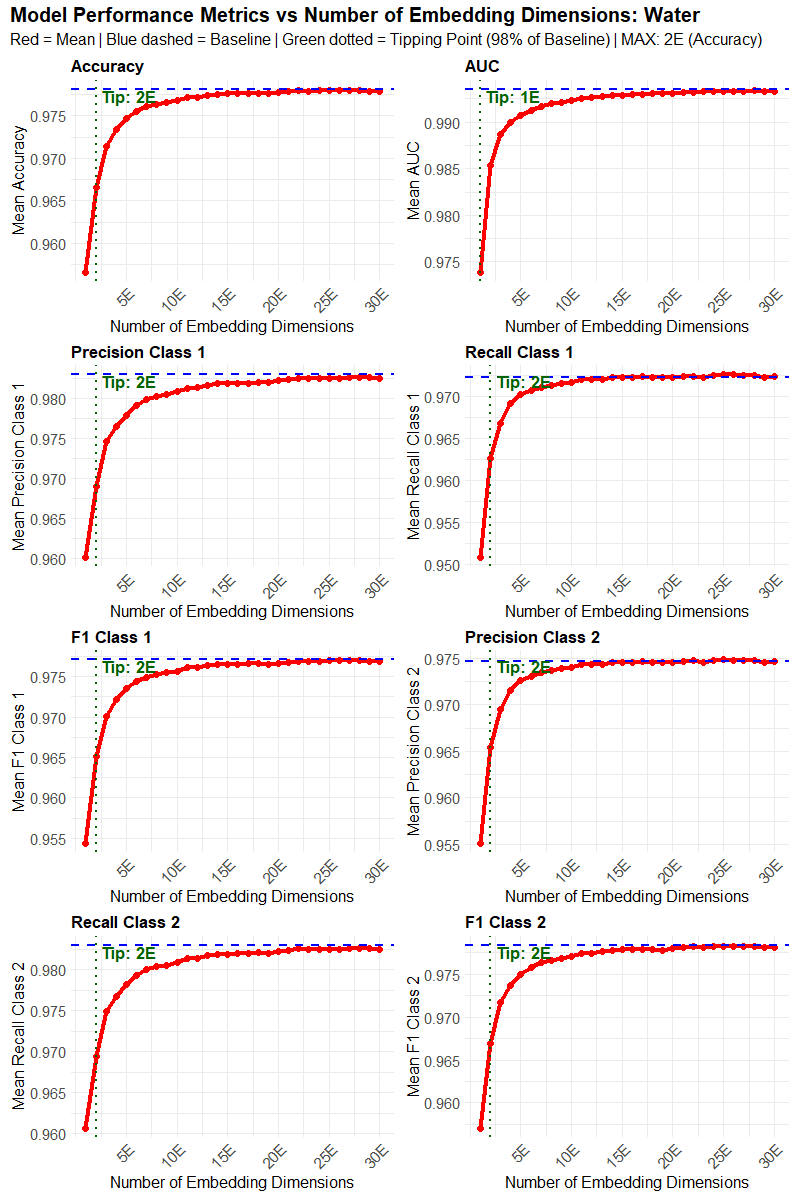}}
        \caption{}
    \end{subfigure}
    \hfill
    \begin{subfigure}{0.45\textwidth}
        \fbox{\includegraphics[width=\linewidth]{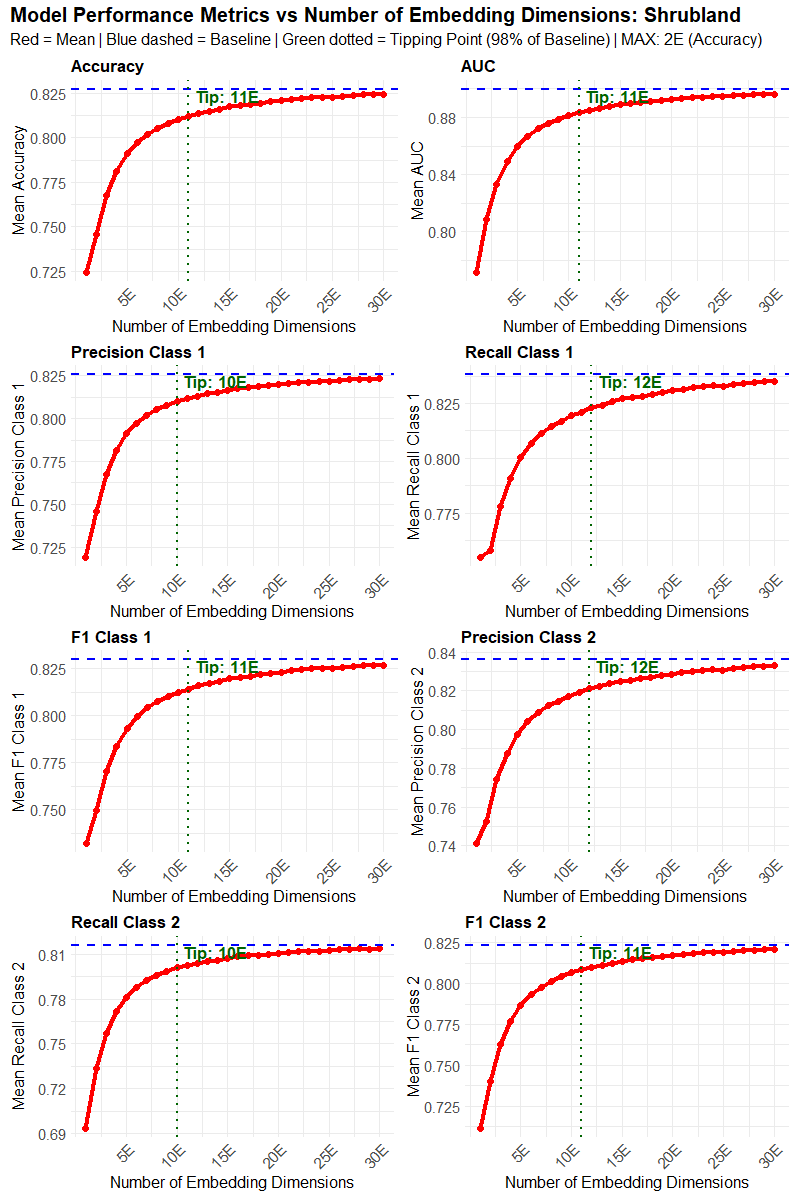}}
        \caption{}
    \end{subfigure}

    \vspace{0.3cm}
    \begin{subfigure}{0.50\textwidth}
        \centering
        \includegraphics[width=\linewidth]{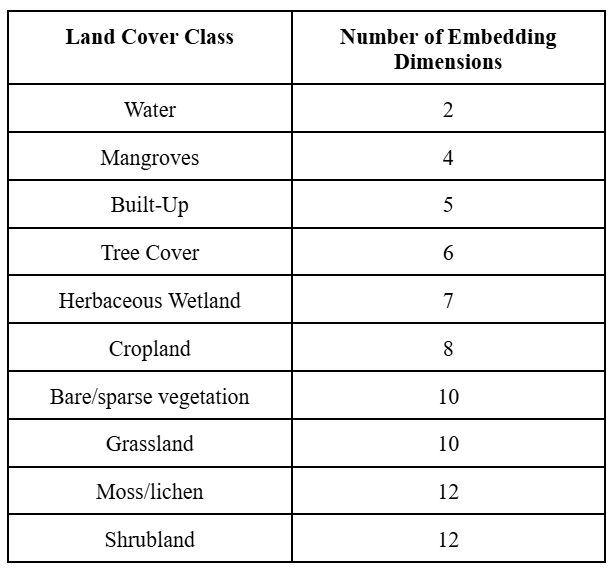}
        \caption{}
    \end{subfigure}

    \caption{Progressive ablation curves and minimum embedding dimensions 
by land cover class. Panels (a) and (b) show eight classification metrics as a function of MDI-ranked embedding dimensions for Water and Shrubland respectively (red = mean across experiments; blue dashed = 64-dimension baseline; green dotted = tipping point at $98\%$ baseline recovery). Water requires as few as 2 dimensions while Shrubland requires up to 12, reflecting differences in spectral 
distinctiveness. Panel (c) summarizes the minimum dimensions required to achieve $98\%$ of baseline performance across all 11 ESA WorldCover 
2020 land cover classes.}
    \label{fig:fig5}
\end{figure}
}

In all cases, performance curves exhibit a characteristic plateau, with rapid gains from the first few dimensions followed by diminishing returns as additional dimensions are added (Supplementary Figure~\ref{fig:s1}). This pattern suggests the existence of considerable redundancy in the embedding space alongside a core of highly informative dimensions that capture distinctive characteristics of each land cover type. The variation in tipping points across classes further indicates that the embedding space encodes class-specific information at different levels of concentration — classes with unique spectral or spatial signatures are captured by fewer, more specialized dimensions, while classes sharing characteristics with others require a broader set. This finding is consistent with the nature of representations learned in foundation models, where information tends to be organized into distributed but non-homogeneous latent structures \citep{Rahman2026}.

Beyond its theoretical implications, this result has direct practical relevance: embedding prioritization enables substantial reductions in computational cost without meaningful loss in classification accuracy. Depending on the land cover class, restricting inference to the minimum 
embedding subset reduces classification time by $20\%$ to $80\%$ relative to the full 64-dimension baseline, offering a pathway toward more efficient deployment of foundation model embeddings in operational contexts.

\subsection{Emergence of highly specialized embedding dimensions}
The analysis of the importance of variables reveals the recurring presence of embedding dimensions with a high contribution to the discrimination of specific coverages. These dimensions show a strong association with a single class, maintaining limited relevance in other categories.

This pattern suggests the existence of highly specialized embedding dimensions that capture distinctive characteristics of certain land cover types. In particular, it is observed that land cover types with clearly distinguishable physical signatures, such as bodies of water or urban areas, tend to have embedding dimensions with a high concentration of importance.

The emergence of these specialized embedding dimensions is consistent with the ability of foundation models to encode physical and environmental properties in high-dimensional latent spaces \citep{Rahman2026}. However, this work does not seek to directly associate embedding dimensions with specific physical variables such as temperature or elevation. Instead, it interprets their behavior functionally, examining how individual dimensions contribute to discriminating between land cover classes and capturing spatial interactions among them.

\footnotesize 

\renewcommand{\arraystretch}{1.0} 
\setlength{\tabcolsep}{5pt} 

\begin{longtable}{
    >{\centering\arraybackslash}p{2.5cm} 
    >{\centering\arraybackslash}p{2.8cm}
    >{\RaggedRight\arraybackslash}p{8.5cm}
}

\caption{Exclusive embedding dimensions and their associated ESA WorldCover 2020 land cover classes, listing dimensions that appear solely in one class's minimum subset alongside a biophysical interpretation of the distinctive spectral, structural, or phenological properties that justify their exclusive association.}
\label{tab:exclusive_embeddings}\\

\toprule
\textbf{Exclusive Embedding Dimension} &
\textbf{Associated Land Cover Class} &
\textbf{Distinctive trait that can justify an exclusive embedding} \\
\midrule
\endfirsthead

\multicolumn{3}{c}{\small\tablename~\thetable{} \textit{(continued from previous page)}}\\
\toprule
\textbf{Exclusive Embedding Dimension} &
\textbf{Associated Land Cover Class} &
\textbf{Distinctive trait that can justify an exclusive embedding} \\
\midrule
\endhead

\midrule
\multicolumn{3}{r}{\small\textit{Continued on next page}}\\
\endfoot

\bottomrule
\endlastfoot

A09, A35
& Built-up
& Urban surfaces and infrastructure: roofs, roads, concrete. Their
radiometric signature is dominated by artificial materials with high SWIR
reflectance compared to NIR and strong Sentinel-1 backscatter. No other
class combines this angular geometry and artificial-material spectrum,
which justifies specific embeddings. \\
\midrule

A12, A50
& Cropland
& Annual crops sown and harvested within a 12-month cycle, with very regular
phenology (well-marked greenness peaks), geometric patterns (plots,
furrows) and often intensive management. Although they can be confused
with grasslands, certain regions of the feature space (e.g.\ very
homogeneous or highly irrigated crops) carry signatures specific to
cropland, explaining their exclusive embeddings. \\
\midrule

A05, A27
& Mangroves
& Evergreen woody vegetation tolerant to salinity in the intertidal zone.
They show dense canopy like the Tree cover class, but with coastal linear
patterns, water mixed between crowns and specific NIR/SWIR signatures due
to permanent waterlogging. This combination (dense forest, brackish water
and coastal context) is unique to mangroves. \\
\midrule

A04, A11, A25, A29
& Herbaceous wetland
& Areas dominated by herbaceous vegetation that is permanently or regularly
flooded (fresh, brackish or saline water). They combine moderate--high
NDVI with very high water indices (NDWI, etc.) and floodplain textures;
no other type combines herbaceous cover and long-lasting flooding in
this way. \\
\midrule

A18, A21, A26
& Shrubland
& Zones dominated by natural shrubs ($\geq$10\% cover), typical of
semi-arid biomes. Although they mix with grassland and bare/sparse
vegetation, subtypes where shrub height, density and spatial pattern
(e.g.\ dense Mediterranean shrublands) generate signatures distinct
enough to occupy an exclusive subspace. \\
\midrule

A16, A23
& Tree cover
& Regions dominated by trees ($\geq$10\% cover), including plantations,
with very high NIR, strong shadows and closed canopy textures. This 3-D
structure and phenological persistence (especially in evergreen forests)
clearly separate them from shrubland and grassland, allowing embeddings
to specialise in representing ``dense forests''. \\
\midrule

A41
& Grassland
& Areas dominated by natural herbaceous vegetation (grasslands, savannas,
prairies) with $\geq$10\% cover and no dominant woody stems. They show
marked seasonality but a homogeneous structure, without long-lasting
flooding (unlike Herbaceous wetland) or intense geometric traces (unlike
Cropland). Embedding A41 likely captures ``well-defined grassland'' far
from crops and shrubs. \\
\midrule

A44, A57
& Moss/lichen
& Surfaces dominated by mosses and/or lichen, typical of tundra and cold
or very arid rocky environments. Their very low structure and spectral
behaviour (moderate NDVI, fine textural patterns over rock or bare soil)
differ markedly from tall herbaceous vegetation and shrubs, justifying
exclusive embeddings for these cryptogamic environments. \\
\midrule

A63
& Bare/sparse vegetation
& Lands with soil, sand or exposed rock and $<$10\% vegetation throughout
the year. This is the ``purest'' bare soil signature: high reflectance in
VIS and SWIR, minimal chlorophyll signal, and very little temporal
variation. No other class maintains such consistently low vegetation
cover year-round, so a dedicated embedding is reasonable. \\
\midrule

A64
& Permanent water bodies
& Permanent water bodies ($\geq$9 months/year), with very strong absorption
in NIR and SWIR, clear linear or polygonal geometry (rivers, lakes) and
low spectral variability. This near-zero NIR/SWIR reflectance and smooth
texture is unique compared to any other class, so a specific embedding
for water is fully expected. \\

\end{longtable}

\subsection{Evidence of shared embedding dimensions between coverages}
Additionally, the analysis shows the existence of embedding dimensions that make significant contributions across multiple coverages. These dimensions are not exclusively associated with one class, but rather participate in the discrimination of different types of coverage. This behavior suggests that certain embedding dimensions capture common patterns or interactions between land cover classes, which is particularly relevant in contexts where classes are not strictly separable. From a geographical perspective, this type of representation is consistent with the existence of transition zones or ecotones, where characteristics of multiple land coverages coexist. From a functional perspective, it is also consistent with land cover classes that share spectral, phenological, or structural characteristics, such as croplands and grasslands, which may exhibit similar seasonal greenness patterns despite differing in management and land use.

The presence of these shared embeddings is consistent with the ability of GFMs to integrate multi-source information and capture complex spatial dynamics \citep{Brown2025}. In this sense, these dimensions cannot be interpreted as exclusive attributes, but rather as components that reflect relationships between classes.

\footnotesize 
\renewcommand{\arraystretch}{1.0}
\setlength{\tabcolsep}{5pt}

\begin{longtable}{
    >{\centering\arraybackslash}p{1.8cm}
    >{\RaggedRight\arraybackslash}p{2cm}
    >{\centering\arraybackslash}p{2cm}
    >{\RaggedRight\arraybackslash}p{2.5cm}
    >{\RaggedRight\arraybackslash}p{6cm}
}

\caption{Shared embedding dimensions and their functional interpretation, grouped by generalist category (low-, mid-, and high-generalist) according to the number of land cover classes in whose minimum subset they appear. For each dimension, the associated ESA WorldCover 2020 classes and a 
common ecological--spectral trait are reported, inferred through the two-stage interpretive process described in Section~2.6.}
\label{tab:generalist_embeddings}\\

\toprule
\textbf{Embedding Dimension} &
\textbf{Associated ESA Land Covers} &
\textbf{Categorization} &
\textbf{Identifier} &
\textbf{Common trait (ecological–spectral synthesis)} \\
\midrule
\endfirsthead

\multicolumn{5}{c}{\small\tablename~\thetable{} \textit{(continued from previous page)}}\\
\toprule
\textbf{Embedding Dimension} &
\textbf{Associated ESA Land Covers} &
\textbf{Categorization} &
\textbf{Identifier} &
\textbf{Common trait (ecological–spectral synthesis)} \\
\midrule
\endhead

\midrule
\multicolumn{5}{r}{\small\textit{Continued on next page}}\\
\endfoot

\bottomrule
\endlastfoot

A07
& Cropland, Moss/lichen
& Low-generalist
& Discontinuous vegetation
& Areas with low and discontinuous vegetation over mineral soil: croplands in fallow or early phenological stages and lichen mats show a soil and sparse vegetation mixture with high visible albedo and marked seasonal variability. \\
\midrule

A08
& Built-up, Moss/lichen
& Low-generalist
& Sparse vegetation over bright surfaces
& Bright, sparsely vegetated substrates such as roofs, concrete and rock with lichens share low chlorophyll signal and rough textures that the model groups as mineral surfaces with minimal vegetation. \\
\midrule

A15
& Moss/lichen, Snow/ice
& Low-generalist
& High latitude and high albedo
& High-latitude or high-altitude cold environments where snow/ice and lichen mats over rock produce bright visible reflectance and spatially homogeneous cryogenic landscapes. \\
\midrule

A28
& Grassland, Tree cover
& Low-generalist
& Dense green vegetation
& Landscapes dominated by green vegetation with strong NIR response where the main difference is canopy density; the embedding captures forest–grassland mosaics with intermediate canopy fraction. \\
\midrule

A34
& Shrubland, Tree cover
& Low-generalist
& Woody perennial
& Woody perennial vegetation with relatively high biomass; shrublands and forests differ in height but share canopy textures and similar spectral and SAR responses. \\
\midrule

A36
& Bare/sparse, Herbaceous wetland
& Low-generalist
& Soil-moisture gradient
& Transition from dry bare soil to wet herbaceous substrates where exposed mud and seasonal wetlands resemble moist bare surfaces, capturing gradients of soil moisture. \\
\midrule

A40
& Grassland, Shrubland
& Low-generalist
& Open low–medium vegetation
& Continuum of open vegetation of low to medium height typical of grasslands and shrublands that share discontinuous structure and similar spectral behaviour. \\
\midrule

A43
& Grassland, Moss/lichen
& Low-generalist
& Low-height vegetation
& Surfaces with creeping or very low vegetation such as sparse grasslands or lichen-dominated tundras, producing moderate NDVI and frequent mixing with exposed soil. \\
\midrule

A53
& Mangroves, Moss/lichen
& Low-generalist
& Extreme conditions
& Vegetation adapted to extreme environments: saline mangroves and cold or arid lichens both show specialized perennial vegetation and complex textures linked to environmental stress. \\
\midrule

A55
& Bare/sparse, Moss/lichen
& Low-generalist
& Rocky sparse vegetation
& Mineral substrates dominated by rock or soil with extremely sparse vegetation where lichen patches barely increase NDVI above bare soil levels. \\
\midrule

A59
& Bare/sparse, Moss/lichen
& Low-generalist
& High-reflectance sparse vegetation
& Reinforces the embedding subspace associated with high-reflectance surfaces with very low vegetation cover where lichens and bare soil are spectrally mixed. \\
\midrule

A10
& Bare/sparse vegetation, Herbaceous wetland, Mangroves
& Mid-generalist
& Soil–flooding gradient
& Gradient from dry bare soil to flooded herbaceous wetlands and mangrove systems, representing strong soil moisture and flooding dynamics across floodplains and coastal zones. \\
\midrule

A13
& Bare/sparse vegetation, Grassland, Shrubland
& Mid-generalist
& Fractional vegetation in drylands
& Semi-arid landscapes characterized by gradual transitions from bare soil to sparse grasslands and shrublands where vegetation fraction varies continuously. \\
\midrule

A14
& Cropland, Grassland, Herbaceous wetland
& Mid-generalist
& Herbaceous vegetation systems
& Land covers dominated by herbaceous vegetation including crops, natural grasslands and wetlands that share strong seasonality and similar canopy spectral signatures. \\
\midrule

A37
& Bare/sparse vegetation, Built-up, Moss/lichen
& Mid-generalist
& Highly reflective sparse vegetation
& Reflective mineral or artificial substrates such as rock, bare soil and urban surfaces with minimal vegetation signals and strong brightness in visible bands. \\
\midrule

A51
& Cropland, Shrubland, Tree cover
& Mid-generalist
& Agroforestry mosaics
& Agricultural landscapes where crops coexist with woody vegetation such as agroforestry systems, plantations or silvopastoral mosaics. \\
\midrule

A52
& Bare/sparse vegetation, Shrubland, Tree cover
& Mid-generalist
& Woody vegetation gradient
& Gradual increase of woody vegetation from nearly bare surfaces to shrublands and open forests representing partial woody cover landscapes. \\
\midrule

A56
& Built-up, Moss/lichen, Snow/ice
& Mid-generalist
& High albedo surfaces
& Surfaces with very high albedo and low chlorophyll signal such as snow, bright urban materials and rocks with lichens. \\
\midrule

A60
& Cropland, Grassland, Shrubland
& Mid-generalist
& Cropland, grassland, shrubland axis
& Rural matrices where agricultural fields, fallows and semi-natural vegetation coexist forming a strong continuum of herbaceous and woody land covers. \\
\midrule

A62
& Grassland, Shrubland, Permanent water bodies
& Mid-generalist
& Riparian mosaics
& Landscapes where open vegetation interacts with water bodies such as riparian corridors and piedmont areas combining vegetation reflectance and water absorption signatures. \\
\midrule

A03
& Cropland, Grassland, Shrubland, Snow/ice
& High-generalist
& Seasonal snow over open vegetation
& Temperate and cold landscapes where open herbaceous vegetation periodically becomes snow covered, generating strong seasonal spectral changes. \\
\midrule

A45
& Bare/sparse vegetation, Grassland, Shrubland, Snow/ice
& High-generalist
& Arid–semiarid environments
& Gradients of sparse vegetation typical of arid or semi-arid regions occasionally influenced by seasonal snow events. \\
\midrule

A61
& Bare/sparse vegetation, Cropland, Moss/lichen, Snow/ice
& High-generalist
& High latitude and mountain mosaics
& Highly heterogeneous landscapes typical of high latitudes or mountains where bare soil, marginal croplands, lichens and snow coexist within extremely seasonal environments. 
\end{longtable}
\normalsize

\subsection{High-generalist embedding dimensions and global behavior} Among the shared dimensions reported in Table~\ref{tab:generalist_embeddings}, high-generalist dimensions (those associated with four or more land cover classes) are distinguished by their consistently low contribution to individual classification tasks. These dimensions have a more homogeneous distribution of importance across coverages and do not show a dominant association with any class.

This behavior suggests that these dimensions capture more general patterns, possibly associated with large-scale environmental gradients or variables that are not decisive for the discrimination of specific land cover types. This type of representation is consistent with the global nature of GAEF embeddings, which seek to provide a consistent description of the state of the Earth's surface on a planetary scale \citep{Houriez2025}. Although these dimensions do not contribute significantly to individual class discrimination, their presence suggests that robust land cover classification depends on the interplay between specialist and shared dimensions, not specialist dimensions alone.

\subsection{Functional organization of the embedding space}
The integration of the observed patterns reveals a functional structure in the embedding space, characterized by the coexistence of different types of dimensions according to their behavior in classification tasks. The results suggest a spectrum of functional roles: from specialist dimensions with high discriminatory capacity for a single land cover class, through low- and mid-generalist dimensions that capture relationships between two or three classes, to high-generalist dimensions that reflect broader environmental patterns.

This organization emerges consistently from empirical analysis, rather than from any explicit structural constraint known to the authors, suggesting that the embedding space encodes structured information related to the geographical organization of the territory. This finding contributes to advancing our understanding of foundation models, by showing that their representations are not arbitrary, but rather reflect patterns that can be interpreted from an ecological and spatial perspective.

\subsection{Visual representation of the universe of embedding dimensions}
\begin{figure}[H]
    \centering
    \fbox{\includegraphics[width=1\linewidth]{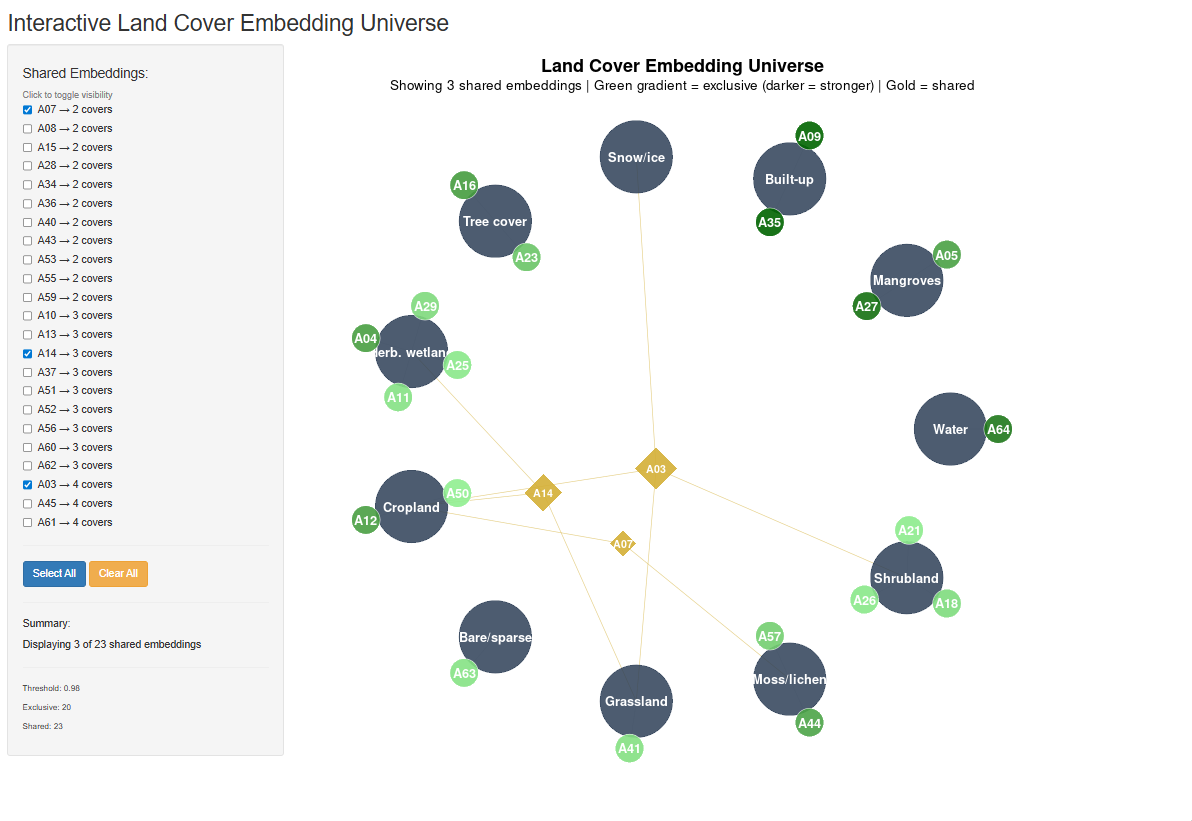}}
    \caption{``Embedding Universe'' visualization at a 98\% baseline performance 
    recovery threshold, generated using the interactive Dashboard 
    \citep{Guthrie2025}. Green nodes represent specialist dimensions 
    (darker = stronger association); gold nodes represent shared dimensions. 
    See Section 2.5 for a full description of the visual 
    encoding. To interact with this plot, access 
    \href{https://alpha-earth-viz.vercel.app/conceptual}{this link}.}
    \label{fig:fig6}
\end{figure}

The embedding universe visualization (Section 2.5) provides an integrated view of the identified patterns. Figure \ref{fig:fig6}, generated using the interactive Dashboard \citep{Guthrie2025}, illustrates the embedding space at a 98\% baseline performance recovery threshold. At this threshold, a predominance of specialist embedding dimensions is observed, with most dimensions positioned in close proximity to a single land cover class, suggesting a high capacity for specialization in the representation space. Shared dimensions, though fewer in number, occupy intermediate positions that highlight inter-class relationships. This visualization provides qualitative evidence of the functional organization described in the preceding analyses.

\subsection{Summary of findings}
Overall, the results obtained do not seek to conclusively validate a hypothesis or to attribute definitive and unique interpretations to individual embedding dimensions, but rather to provide empirical evidence in support of a functional interpretability framework. The observation of 
consistent patterns in the importance of embeddings, their differential behavior between land cover classes, and their organization in the representation space suggests that it is possible to characterize these dimensions in functional terms --- while acknowledging that the interpretations proposed here are task-dependent, biophysically informed approximations rather than fixed semantic labels.

These findings motivate the conceptualization of the embedding space as a structured system, in which dimensions can be interpreted according to their role in representing land cover and its interactions. In this sense, the results constitute the empirical basis for the proposed interpretability framework, which reveals a hierarchical functional structure within the embedding space --- one that, once characterized, provides a transferable analytical chassis through which embeddings can be interpreted in relation to any geospatial domain or dataset, beyond the land cover context explored here.

\section{Discussion}

\subsection{Interpretability in GFMs: from physical variables to functional structures}
GFMs have demonstrated significant progress in Earth observation tasks, thanks to their ability to integrate multi-source data into latent representations of high dimensionality \citep{Brown2025, Houriez2025}. However, this progress has been accompanied by a fundamental challenge: the lack of interpretability of embeddings, which encode environmental information into abstract vectors with no direct correspondence to observable physical variables \citep{Rahman2026}.

Existing approaches have addressed this limitation primarily by associating embedding dimensions with continuous variables, such as temperature, precipitation, or elevation, seeking to identify explicit physical relationships within the latent space \citep{Rahman2026}. While these approaches represent an important advance, they focus on a physical-variable interpretation, which does not necessarily capture the categorical organization of the territory in terms of land cover.

In contrast, this paper proposes a complementary approach based on functional interpretability, in which embeddings are interpreted according to their role in discriminating between coverages and representing their interactions. This approach allows the latent space to be connected to fundamental geographic concepts, such as coverage units and transition zones, extending the scope of interpretability beyond traditional physical variables.

\subsection{Functional organization of the embedding space}
The results suggest that the embedding space has a non-uniform structure, where dimensions play different roles. This organization can be interpreted as a functional spectrum, from specialist dimensions that capture the distinctive characteristics of a single land cover class, through low- and mid-generalist dimensions that encode relationships between two or three classes, to high-generalist dimensions that reflect broader environmental patterns.

Notably, specialist dimensions, which concentrate importance on unique spectral-structural signatures like ``Water'' or ``Built-up'', effectively represent the ``core'' of land-cover units. Meanwhile, the identification of mid-generalist dimensions suggests that the embedding space captures the fuzzy logic of environmental transitions. These mid-generalist dimensions act as semantic bridges, representing ecotones where classes overlap --- such as the moisture-driven gradient between 'Bare soil' and 'Herbaceous wetlands'. This functional organization implies that foundation models do not merely memorize pixels but encode a structured blueprint of environmental perception.

This functional structure emerges consistently from empirical analysis, suggesting that embeddings are not arbitrary representations but rather reflect regularities inherent in the organization of the territory. In this sense, the embedding space can be understood as a system in which different dimensions capture different levels of abstraction of geographical reality.

This finding is consistent with evidence that foundation models are capable of learning rich and generalizable representations from heterogeneous data \citep{Brown2025}, but it offers a new perspective by suggesting that these representations also possess a structure that can be interpreted in terms of discrete spatial categories and their interactions.

\subsection{Specialist embeddings and coverage core representation}
The identification of embeddings with high specialization in certain coverages suggests that the latent space contains dimensions that capture distinctive characteristics of specific categories of the territory. This behavior is particularly evident in coverages with well-defined physical signatures, where the separability between classes is clearer.

From an interpretative perspective, these embeddings can be understood as representations of the ``cores'' of land cover classes, i.e., regions where the characteristics of a class are dominant and relatively homogeneous. In these contexts, specialized embeddings enable accurate discrimination, acting as robust indicators of the presence of a land cover class.

This type of representation is consistent with the ability of foundation models to capture relevant physical properties of the environment, although in this case the interpretation is performed in terms of geographic categories rather than continuous variables \citep{Rahman2026}.

\subsection{Low- and mid-generalist dimensions and ecotone representation}
One of the most significant contributions of this work is the identification of generalist embedding dimensions associated with a limited number of land cover classes. These dimensions serve a dual role: they capture characteristics common to classes that overlap spatially or functionally, as in transition zones or ecotones where coverages lack clearly defined boundaries, and they complement specialist dimensions in the classification of individual land cover classes where no single dimension is sufficient on its own.

The existence of these dimensions suggests that foundation models not only capture discrete patterns, but also the interactions between them, integrating information from multiple sources and spatial scales. This behavior is consistent with the nature of environmental systems, where transitions between land covers are frequent and play a key role in the dynamics of the territory. Their association with two or three specific classes makes low- and mid-generalist dimensions particularly suited to representing the inter-class relationships characteristic of ecotonal boundaries.

In this sense, these dimensions constitute a bridge between the continuous representation of latent space and the discrete structure of coverages, allowing the complexity inherent in geographic systems to be captured. This aspect has been little explored in the literature, despite its relevance for applications such as monitoring land use changes or detecting degradation processes.

\subsection{High-generalist dimensions and environmental gradient representation}
In addition to specialist, low-, and mid-generalist dimensions, high-generalist dimensions are identified that exhibit more general behavior, without a dominant association with any individual land cover class. These dimensions can be interpreted as representations of large-scale environmental gradients, such as climatic or topographic conditions.

This type of dimension reflects the global nature of GFMs, which seek to provide a consistent representation of the planet across different regions and contexts \citep{Houriez2025}. Although these dimensions are not highly discriminative in classification tasks, their presence indicates that the embedding space incorporates information relevant to the general characterization of the environment.

From an interpretive perspective, high-generalist dimensions can be considered as a higher level of abstraction, capturing properties of the system that transcend the discrete categories of land cover.

\subsection{Uninterpreted dimensions and their significance}
Of the 64 embedding dimensions in GAEF’s latent vector space, the preceding analyses yield functional interpretations for 43. The remaining 21 dimensions were not found within the minimum embedding subset for any land cover class across more than 130,000 independent experiments. These uninterpreted dimensions may encode information outside the land cover domain --- such as broader environmental gradients, atmospheric conditions, or properties of the built environment --- or may represent aspects of the Earth’s surface that the current experimental framework, focused on land cover classification, is not designed to capture.

This constitutes an important direction for future research. Cross-domain experiments targeting alternative domains (e.g., infrastructure, climate, or land use) could reveal whether these dimensions carry structured information that becomes legible under a different contextual lens. Additionally, targeted validation experiments using other datasets may help determine whether these dimensions truly encode subtle environmental signals that fall below the  discrimination threshold in the current design.

\subsection{Implications for interpretability and remote sensing}
The proposed framework has important implications for the interpretability of foundation models in remote sensing. First, it allows interpretation to shift from a variable-based approach to a function-based approach, which is more suitable for analyzing complex geographic phenomena.

Secondly, identifying different types of embedding dimensions opens up the possibility of designing more efficient and explainable models by selecting subsets of dimensions according to their function. For example, specialist dimensions could be used to improve accuracy in class-specific identification, while low- and mid-generalist dimensions could be key to improving change detection or edge delimitation.

Thirdly, this approach allows the interpretation of models to be integrated with concepts specific to geography and ecology, facilitating their use in scientific and decision-making contexts.

\subsection{Practical implications}
The findings of this study transcend theoretical interpretability, offering immediate practical benefits for the deployment of GFMs in large-scale operational environments:
\begin{itemize}
	\item Computational Cost Reduction: By identifying that accurate land-cover classification ($\geq$ 98\% of baseline) requires as few as 2 to 12 dimensions, organizations can implement a dimension pruning strategy. This reduces the data footprint of high-resolution (10 m) global mosaics by up to 90\%, significantly lowering storage costs and memory overhead during inference.
    
	\item Targeted Feature Selection: Instead of utilizing the full 64-dimensional embedding as a ``black box,'' practitioners can now select specialist dimensions tailored to specific domains. For instance, urban monitoring tasks can prioritize dimensions A09 and A35, while coastal management projects can focus on mangrove-specific dimensions like A05 and A27.
    
	\item Explainable AI (XAI) for Decision Support: The functional taxonomy provides a ``semantic bridge'' for non-expert stakeholders. By utilizing the ``Embedding Universe'' visualization, model predictions move from abstract vectors to justifiable geographic associations, facilitating trust in AI-driven climate adaptation and infrastructure planning.

    \item Enhanced Change Detection: Identifying shared dimensions that represent ecotones allows for more sensitive monitoring of ecological transition zones. These dimensions can be used as early-warning indicators for land degradation or deforestation, where subtle shifts in the latent space signal transitions before they are evident in categorical labels.
\end{itemize}

\subsection{Limitations of the study}
Despite the contributions presented, this work has several limitations that should be considered. First, the analysis is based on importance measures derived from machine learning models, which may be influenced by the algorithm used and the distribution of the data. This implies that the interpretation of the embeddings depends, to some extent, on the experimental context.

Secondly, the functional classification of embedding dimensions is derived from progressive ablation at a 98\% baseline performance recovery threshold. While this threshold is grounded in classification performance rather than arbitrary cutoffs, different recovery thresholds would yield different minimum subsets and potentially different functional classifications. The sensitivity of results to this choice warrants further analysis.

Additionally, although the proposed visualization facilitates interpretation, it represents a simplification of the embedding space, which is actually high-dimensional. Therefore, the relationships observed should be understood as an approximation of the actual structure of the system.

Furthermore, although this study captures regions of interest distributed globally, the functional roles identified for each embedding dimension may shift when the framework is applied at a regional scale. Classification accuracy itself varies geographically (Supplementary Figure~\ref{fig:s2}), suggesting that regional land cover conditions influence the discriminatory capacity of the embedding space. In geographically constrained contexts, reduced land cover diversity and environmental variability could diminish the discriminatory capacity of dimensions that are informative at a global level, effectively flattening the importance distribution.

Finally, this study does not address the validation of the proposed interpretations in terms of specific physical variables or environmental processes, which constitutes a line of future work.

\subsection{Future lines of research}
The results obtained open up multiple lines of research. One relevant direction is the validation of the proposed framework by associating embeddings with specific physical variables, which would allow the integration of functional and physical interpretability approaches.

Another line of research consists of analyzing the stability of the functional classification of embeddings in different regions and temporal scales, evaluating their capacity for generalization. Likewise, the study of the temporal dynamics of embeddings could provide information on processes of change in the territory.

A third line of research revolves around geographic variation. The geographic variation in classification performance revealed across the global experiments (Supplementary Figure~\ref{fig:s2}) suggests that the functional roles of individual embedding dimensions may not be uniform across biomes or climatic zones. A systematic investigation of the geographic determinants of embedding discriminatory capacity constitutes a promising direction for future research.

Additionally, the proposed framework could be extended to other remote sensing tasks, such as change detection, semantic segmentation, or environmental variable prediction, exploring the role of different types of embeddings in each context.

Finally, integrating these results into interactive analysis and decision-making systems --- such as climate adaptation planning tools and environmental monitoring platforms --- represents an opportunity to bring foundation models closer to practical applications, facilitating their interpretation by non-specialist users.

\section{Conclusion}
This paper presents a functional interpretability framework for GFMs, aimed at understanding the internal structure of embeddings in relation to the organization of territory. Based on an exploratory analysis involving massive experimentation and structural characterization of the embedding space, the evidence suggests that these representations are not arbitrary, but rather present consistent patterns that allow their behavior to be interpreted in terms of land cover and their interactions.

The results suggest that embedding space can be understood as a functionally organized system in which dimensions with different roles coexist: specialist dimensions highly specialized in representing specific land cover classes, low- and mid-generalist dimensions associated with relationships between a limited number of classes, and high-generalist dimensions related to broader environmental gradients. This hierarchical organization allows the latent space of the foundation models to be connected to fundamental geographic concepts, such as land cover units and ecotones, offering an alternative perspective to interpretability approaches based exclusively on physical variables.

In this sense, the main contribution of this work lies in proposing an interpretability approach that shifts the focus from identifying variables to understanding functions within the embedding space. This approach opens up new possibilities for the analysis of foundation models in remote sensing, facilitating their use in scientific contexts and in informed decision-making.

Finally, this study lays the groundwork for future research aimed at validating and extending the proposed framework, including integration with biophysical variables, spatial and temporal stability analysis, and its application in different Earth observation tasks. Together, these advances will help bridge the gap between the high predictive performance of foundation models and their scientific interpretation, promoting a more transparent and explainable use of these technologies.


\section{Statements and Declarations}

\subsection{Competing Interests}
The authors have no competing interests to declare that are relevant to the content of this article.

\subsection{Clinical trial number}
Not applicable.

\subsection{Funding}
The authors did not receive support from any organization for the submitted work.

\subsection{Ethical responsibilities of Authors}
All authors have read, understood, and have complied as applicable with the statement on “Ethical Responsibilities of Authors” as found in the Instructions for Authors.

\subsection{Data Availability Statement}
No new primary data were created in this study. The analysis was conducted using publicly available remote sensing data from the Copernicus Open Access Hub. The derived data products supporting the findings of this study are available from the corresponding author upon request.

\subsection{Code Availability Statement}
The source code for the ``What on Earth is AlphaEarth?'' interactive Dashboard 
is available at \cite{Guthrie2025_repo}. The experimental notebook used to 
conduct the classification experiments is available at \cite{BenavidesMz2025}.

\section{Acknowledgments}
\begin{itemize}
    \item I.F.B would like to thank his colleagues at the Roux Institute (The Institute for Experiential Artificial Intelligence and the Artificial Intelligence for Climate and Sustainability Groups) as well as the Gulf of Maine Research Institute.
    \item J.G. would like to thank his family; his colleagues at the Sustainability and Data Sciences Lab and at Enodia, Inc.; and the Artificial Intelligence for Climate and Sustainability Group and the Institute for Experiential Artificial Intelligence for their encouragement and support.
    \item J.E.A. would like to thank his colleagues at Universidad Católica de Manizales.
    \item Y.A.G. would like to thank to Universidad Católica de Manizales.
    \item A.I.G. would like to thank her colleagues at Universidad Nacional de Colombia, Palmira Campus.
    \item C.V.P. would like to thank his colleagues at Universidad Nacional de Colombia, Palmira Campus.
\end{itemize}

\pagebreak
\bibliographystyle{apalike}
\bibliography{embeddings}


\clearpage
\newpage
\setcounter{figure}{0}
\renewcommand{\thefigure}{\arabic{figure}}
\renewcommand{\figurename}{\textbf{Supplementary Figure}}
\captionsetup[figure]{labelfont=bf}

\section*{Supplementary Information}

\begin{figure}[h]
    \centering
    \fbox{\includegraphics[scale=0.45]{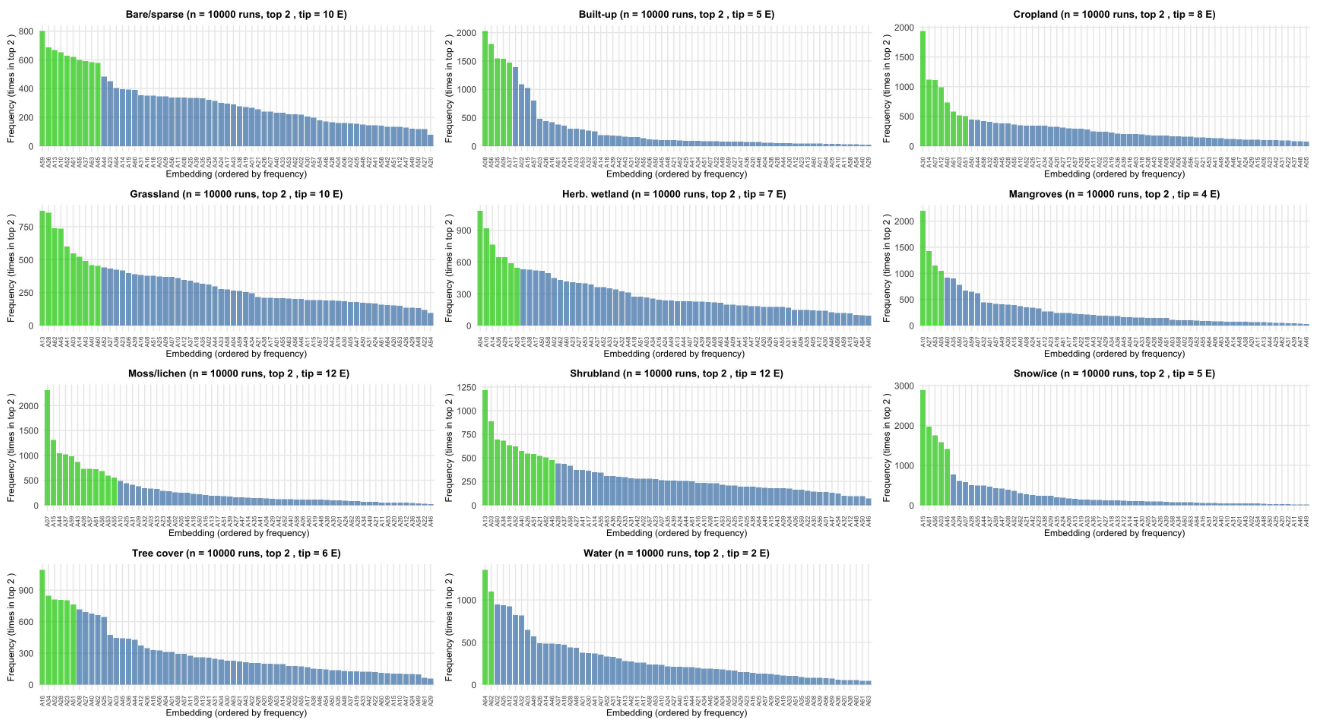}}
    \caption{\textbf{Embedding frequency distributions by land cover class.} For each class, embedding dimensions are ranked by frequency of appearance in the top 2 most important positions across 10,000 experimental runs. Green bars indicate dimensions within the tipping point threshold; blue bars indicate remaining dimensions. Generated using the interactive Dashboard \citep{Guthrie2025}.}
    \label{fig:s1}
\end{figure}

\begin{figure}[h]
    \centering
    \fbox{\includegraphics[scale=0.4]{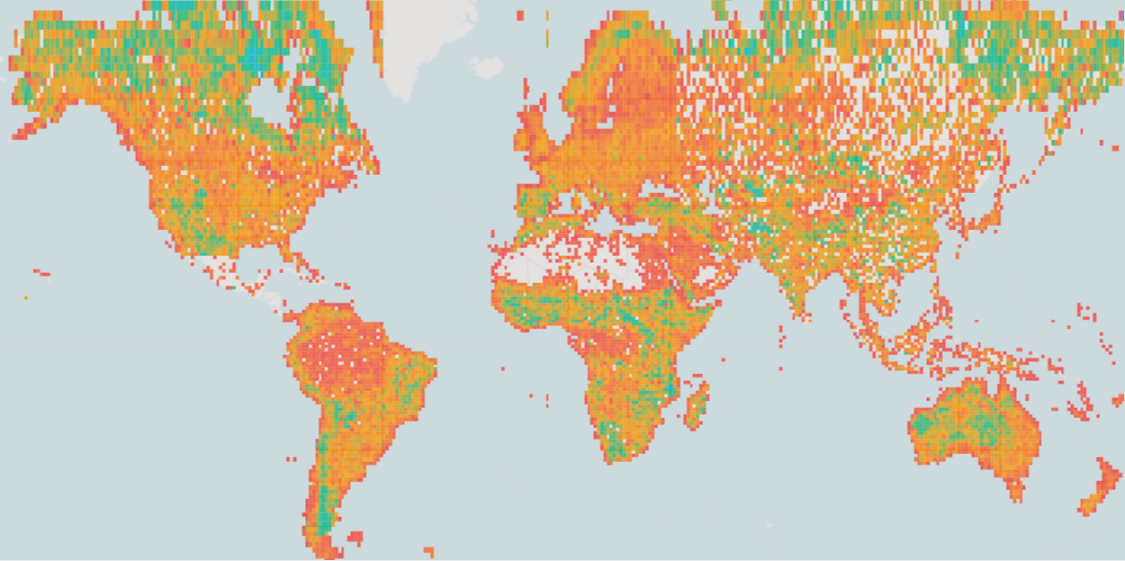}}
    \caption{\textbf{Grid-based geographic performance heatmap of classification accuracy across global experiments.} Each cell represents the mean classification accuracy aggregated across all experiments conducted within that geographic grid cell. Red cells indicate higher accuracy (>90\%), orange cells indicate moderate accuracy (80–90\%), and blue cells indicate lower accuracy (<80\%). Generated using the interactive Dashboard \citep{Guthrie2025}.}
    \label{fig:s2}
\end{figure}

\end{document}